\documentclass{article}

\usepackage{fullpage}

\usepackage[utf8]{inputenc} % allow utf-8 input
\usepackage[T1]{fontenc}    % use 8-bit T1 fonts
\usepackage{hyperref}       % hyperlinks
\usepackage{url}            % simple URL typesetting
\usepackage{booktabs}       % professional-quality tables
\usepackage{amsfonts}       % blackboard math symbols
\usepackage{nicefrac}       % compact symbols for 1/2, etc.
\usepackage{microtype}      % microtypography
\usepackage{graphicx}
\usepackage{wrapfig}

\newtheorem{definition}{Definition}

\title{Appearance invariance in convolutional networks with neighborhood similarity}

\author{
  Tolga Tasdizen \and Mehdi Sajjadi \and Mehran Javanmardi \and Nisha Ramesh\\
    \\
  Scientific Computing and Imaging Institute\\
  University of Utah\\
  Salt Lake City, UT 84106 \\
  \texttt{tolga@sci.utah.edu} \\
}

\begin{document}

\maketitle

\begin{abstract}
We present a neighborhood similarity layer (NSL) which induces appearance invariance in a network when used in conjunction with convolutional layers. We are motivated by the observation that, even though convolutional networks have low generalization error, their generalization capability does not extend to samples which are not represented by the training data. For instance, while novel appearances of learned concepts pose no problem for the human visual system, feedforward convolutional networks are generally not successful in such situations. Motivated by the Gestalt principle of grouping with respect to similarity, the proposed NSL transforms its input feature map using the feature vectors at each pixel as a frame of reference, i.e. center of attention, for its surrounding neighborhood. This transformation is spatially varying, hence not a convolution. It is differentiable; therefore, networks including the proposed layer can be trained in an end-to-end manner. We analyze the invariance of NSL to significant changes in appearance that are not represented in the training data. We also demonstrate its advantages for digit recognition, semantic labeling and cell detection problems. 
 \end{abstract}

\section{Introduction}

The recent successes of deep learning are partially attributed to supervised training of networks with large numbers of parameters using large datasets. In computer vision, supervised training of convolutional networks with very large labeled datasets provide state-of-the-art solutions in many applications such as object recognition, image captioning and question answering. While it has been shown that convolutional networks have low generalization error, their generalization capability does not extend to samples which are not adequately represented by the training data. A potential source of mismatch between the training data distribution and new samples is appearance. To a human, the images shown in Figure~\ref{fig:maps} (top row) unambiguously represent the digits "4", "2" and "6" whereas a convolutional network trained on the original MNIST dataset has a low probability of producing the correct answer for the modified digit images. The reason that a human has an easy time at this task is not because he has previously been exposed to the particular representations of the digits shown in Figure~\ref{fig:maps}, but because he is able to adapt to novel appearances of learned concepts. Invariances to a predetermined set of transformations such as translation, rotation, contrast and noise can be taught to the network via methods such as tangent prop~\cite{Simard1991} and data augmentation~\cite{Goodfellow-et-al-2016}; however, these methods can not adapt to new appearances such as those shown in Figure~\ref{fig:maps}. Similarly, domain adaptation~\cite{Chen2011, Ganin2016} offers a solution only if a sufficient number of images in the target domain are available.

We propose a novel neighborhood similarity layer (NSL) and show that when used in a convolutional network it can greatly increase its generalization accuracy to novel appearances without requiring domain adaptation. The NSL is motivated by the Gestalt principle of grouping with respect to similarity. More specifically, NSL computes normalized inner products of feature vectors of the previous layer between a central pixel which acts as a frame of reference and the spatial neighborhood of that pixel. NSL is a parameter-free layer, which if used after the early convolutional layers in a network, can be seen as inducing appearance invariance. We show that the inclusion of the NSL results in good generalization accuracy for classification and semantic labeling, without using domain adaptation, when the source and target domains differ. Remarkably, the accuracy of the network with NSL trained on MNIST~\cite{Lecun1998} and tested on SVHN~\cite{Netzer2011}, without any domain adaptation, surpasses state-of-the-art domain adaptation results. Cross-domain appearance invariance without domain adaptation is a step towards reasoning about novel representations of learned concepts. We also show that these results can be further improved when NSL is used in conjunction with domain adaptation. Finally, we demonstrate in the context of cell detection, that NSL can improve accuracy by introducing contrast invariance when the source and target domains are the same. 

\section{Related Work}
Invariances to a predetermined set of transformations can be learned during training. Tangent prop~\cite{Simard1991} penalizes the derivative of the network's output with respect to directions corresponding to the desired invariances such as rotation and scale change. Data augmentation~\cite{Goodfellow-et-al-2016} can be seen as a non-infinitesimal version of tangent prop where new labeled samples are generated with the chosen transformations and used in directly in supervised training. Type-invariant pooling~\cite{Laptev2016} creates a more compact decision layer than data augmentation by creating transformation invariant features rather than capturing all possible transformations. Data augmentation has also been applied to unlabeled samples by penalizing the differences of the network's output to transformed versions of an unlabeled sample~\cite{sajjadi2016regularization}. Contractive learning~\cite{rifai2011contractive} and double backprop~\cite{Drucker1992} penalize the magnitude of the derivative of the network's output. This corresponds to promoting an output that varies slowly with respect to any change in the input such as additive noise. Adversarial training~\cite{Szegedy2014,Goodfellow2015} can be seen as a non-infinitesimal version of contractive learning. Scattering convolution networks~\cite{Bruna2013} compute translation invariant descriptors that are also stable under deformations. 
Maxout networks~\cite{Goodfellow2013} can also learn invariances present in the training data. However, these methods can not adapt to variations such as novel appearances that are not predefined or represented in the training data.

Unsupervised domain adaptation makes use of labeled data in a source domain and unlabeled data in the target domain. Co-training~\cite{Blum1998} which uses classifiers with different views to generate pseudo-labels has been used for domain adaptation~\cite{Chen2011}. High confidence pseudo-labels are slowly added as labeled samples in the target domain. Tri-training~\cite{Saito2017} uses three classifiers: two for generating the pseudo-labels and one exclusively for the target domain. The advantage of tri-training over co-training is that it does not need partitioning the input features to two different views. Other methods use an unsupervised reconstruction task in the target domain with a supervised classification task in the source domain~\cite{Ghifary2016}. Residual transfer networks ease the assumption of a shared classifier between the source and target domains~\cite{Long2016}. Domain adversarial networks~\cite{Ganin2016} use a gradient reversal layer to extract domain invariant features for classification. Features are domain invariant if a domain classifier can not predict which domain the input that gave rise to the features came from. In~\cite{Long2015,Tzeng2014}, the maximization of the domain classification loss between source and train domain features is replaced with the minimization of the maximum mean discrepancy. In domain separation networks~\cite{Bousmalis2016a}, the private features of each domain is extracted separately from the common features. The common component of the features is further used for classification purposes on both domains while private and common parts are used for reconstruction. In a recent work~\cite{Bousmalis2016b}, domain invariance is not forced on the feature representation, rather it is recommended to adapt the source images to the target domain using generative adversarial networks~\cite{Goodfellow2014} and perform classification in the target domain only. Domain adaptation is possible only if the target domain is represented by a sufficiently large set of samples, but can not be used for a novel representation of a learned concept without an accompanying target domain dataset.

Similarity between square patches of image intensities is used to define weights for the purpose of denoising in the non-local means algorithm~\cite{Buades2005}. Non-local means uses a Gaussian kernel on the Euclidean distance directly in the space of image intensities. While the proposed NSL can also be applied directly to input image values, we formulate it as a layer that can be used after any image feature map generating layer in a network. Explicit and global perceptual grouping with iterative inference for deep fully connected networks was investigated in~\cite{Greff2016}. NSL performs implicit and local perceptual grouping in a feedforward convolutional network. Finally, using a kernel mapping between pairs of pixels in image, as introduced in convolutional kernel networks (CKN)~\cite{Mairal2014}, is related to the proposed NSL. However, CKNs learn a fixed set of convolutional masks from images in an unsupervised~\cite{Mairal2014} and supervised~\cite{Mairal2016} way to approximate the kernel mapping which we argue leads to low quality approximations for images with appearances that are not represented by the training data. Furthermore, the choice of a Gaussian kernel in~\cite{Mairal2014,Mairal2016} requires learning a standard deviation which further ties the mapping to a particular set of images. 

\section{Neighborhood Similarity Layer}
\begin{wrapfigure}{r}{0.37\textwidth}
\begin{center}
%\vspace{-0.5cm}
\includegraphics[width=0.35\textwidth]{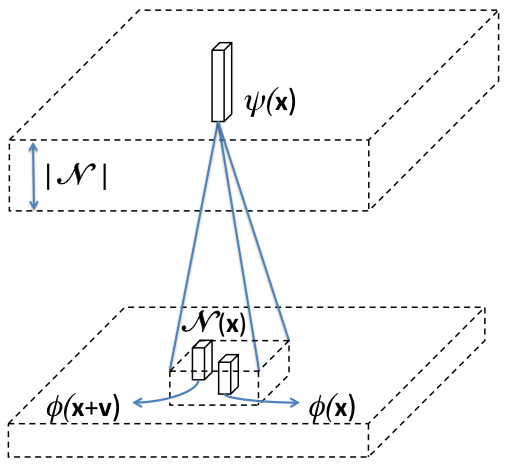}
%\vspace{-0.5cm}
\end{center}
\caption{NSL transforms a feature map $\phi$ to a similarity map $\psi$. $\phi(\mathbf{x})$ acts as a frame of reference, i.e. center of attention, to transform $\phi(\mathbf{x}+\mathbf{v})$.}
\label{fig:nsl}
%\vspace{-0.3cm}
\end{wrapfigure}
The NSL is motivated by the Gestalt principle of similarity, i.e. structures with similar appearance are grouped and processed together in perception. From the perspective of network design, when used after the initial convolutional layers, NSL is an appearance invariance inducing transformation on image feature maps. It transforms its input feature map using the feature vector at each pixel as a frame of reference, i.e. center of attention, for its surrounding neighborhood. To formulate NSL more precisely, we first define image feature maps and neighborhood structures:
\begin{definition}
An image feature map is a n-dimensional vector valued function $\phi:\Omega\rightarrow \mathbb{R}^n$ where $\Omega$ is the discrete d-dimensional image grid and $\mathbb{R}^n$ is the n-dimensional Euclidean space. 
\end{definition}
\begin{definition}
A neighborhood structure ${\cal N}$ is a set of $m$ d-dimensional, non-zero offset vectors $\mathbf{v}\in\Omega$. The neighborhood of a pixel $\mathbf{x}\in\Omega$ then follows as the ordered set ${\cal N}(\mathbf{x}) = \left\{\mathbf{x}+\mathbf{v}:\mathbf{v}\in{\cal N}\right\}$. 
\end{definition}
In this paper, ${\cal N}(\mathbf{x})$ is taken as a square patch around $\mathbf{x}$ excluding the center pixel $\mathbf{x}$ as shown in Figure~\ref{fig:nsl}. The NSL transformation of the feature map  $\phi$ to similarity map is formulated as: 
\begin{definition}
Given a feature map $\phi:\Omega\rightarrow  \mathbb{R}^n$ and a neighborhood structure ${\cal N}$, the neighborhood similarity layer $\psi: \Omega \rightarrow \mathbb{R}^m$ is the vector map of normalized inner products
\begin{equation}
\psi(\mathbf{x},\phi,{\cal N}) = 
\left[
\frac{\left\langle\phi(\mathbf{x}+\mathbf{v})-\bar{\phi},\phi(\mathbf{x})-\bar{\phi}\right\rangle}
{\left(\left\langle\phi(\mathbf{x}+\mathbf{v})-\bar{\phi},\phi(\mathbf{x}+\mathbf{v})-\bar{\phi}\right\rangle
\left\langle\phi(\mathbf{x})-\bar{\phi},\phi(\mathbf{x})-\bar{\phi}\right\rangle\right)^{1/2}}
\right]_{\mathbf{v}\in{\cal N}}
\label{eqn:nsl}
\end{equation}
where $\bar{\phi} =(1/{|\Omega|})\sum_{\mathbf{x}\in\Omega} \phi(\mathbf{x})$ is the mean feature vector over $\Omega$.
\end{definition}

\begin{figure}[htb]
\begin{center}
%\vspace{-0.5cm}
\includegraphics[width=0.98\textwidth]{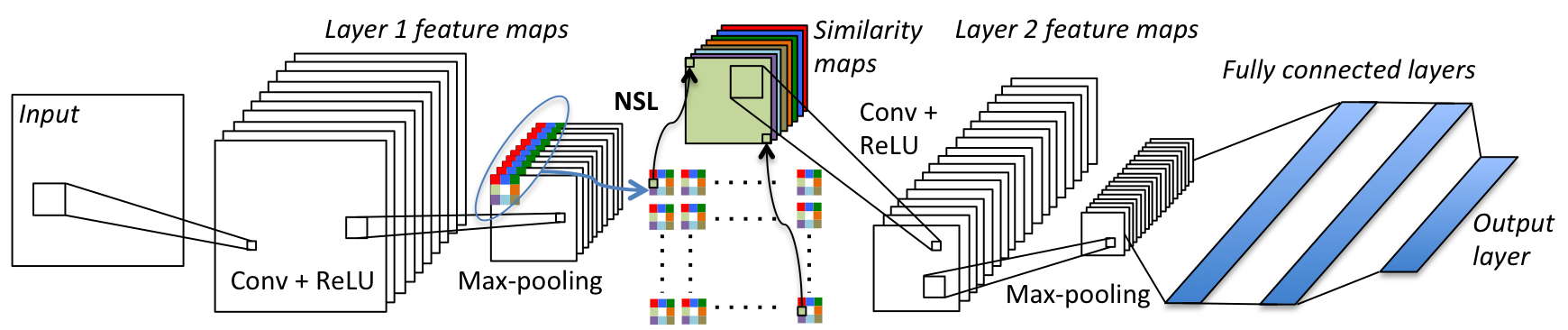}
%\vspace{-0.2cm}
\end{center}
\caption{NSL placed between two layers of a network. At any location $\mathbf{x}$ (top left corner shown), the NSL computes a neighborhood of similarities to $\phi(\mathbf{x})$. This creates an image of two dimensional neighborhood similarities which are flattened into one dimensional vectors creating the output of the NSL. The relationship between a $3\times 3$ ${\cal N}$ and 8 similarity maps are shown with the color coding.}
\label{fig:arch}
%\vspace{-0.7cm}
\end{figure}

We emphasize the following properties of NSL: (i) it is a parameter free layer and hence independent of training data, (ii) it is not a convolution, but a spatially varying operation that uses the feature vectors at each pixel as a frame of reference, (iii) the output feature map $\psi$ has the same dimensionality as the number of pixels in ${\cal N}$ and (iv) when ${\cal N}$ is a square patch, the vector $\psi(\mathbf{x})$ corresponds to a square patch of similarities around and excluding $\mathbf{x}$. A NSL can be placed after any feature map producing layer in a network as shown in Figure~\ref{fig:arch}. Note that, a convolutional layer following a NSL operates on a map of feature vectors which correspond to square patches of similarity vectors. Any number of NSL may be used as different layers of a network. In this paper, our experiments are focused on using a single NSL after the first convolutional layer of networks as an appearance invariance inducing transformation. 
Even though NSL (\ref{eqn:nsl}) is a parameter-free layer, we need to compute its gradient to enable the backpropagation of errors through a NSL when it is used in a network. Letting $\tilde{\phi}=\phi-\bar{\phi}$, it follows from (\ref{eqn:nsl}) that 
\begin{equation}
\frac{\partial \psi}{\partial \phi} =  \frac{\partial \tilde{\phi}}{\partial \phi}\frac{\partial \psi}{\partial \tilde{\phi}} =
\left(1-\frac{1}{|\Omega|}\right)
\left[
\frac{\tilde{\phi}(\mathbf{x}+\mathbf{v})}{||\tilde{\phi}(\mathbf{x}+\mathbf{v})||\cdot||\tilde{\phi}(\mathbf{x})||}-
\frac{\left\langle\tilde{\phi}(\mathbf{x}+\mathbf{v}),\tilde{\phi}(\mathbf{x})\right\rangle\tilde{\phi}(\mathbf{x})}
{||\tilde{\phi}(\mathbf{x}+\mathbf{v})||\cdot||\tilde{\phi}(\mathbf{x})||^3}
\right]_{\mathbf{v}\in{\cal N}}.
\label{eqn:nslderivative1}
\end{equation}

\section{Appearance Invariance Properties of the Neighborhood Similarity Layer}
\label{sec:invariance}
Consider the case where the input image consists of a foreground object and background. Let $p_f(\phi)$ and $p_b(\phi)$ denote the conditional probability density functions in $\mathbb{R}^n$ for the foreground and background, respectively. Let $P_f$ and $P_b$ denote the {\em apriori} probability of belonging to the foreground and background, respectively. Then, the probability density function for $\phi$ is given as $p(\phi) = P_f p_f(\phi) + P_b p_b(\phi)$. Let $\mu_f$ and $\mu_b$ denote the conditional expectation of $\phi$ over $p_f(\phi)$ and $p_b(\phi)$, respectively. The expected value of $\phi$ over $p(\phi)$ is found as $\mu = P_f\mu_f + P_b\mu_b$. NSL centers $\phi$ by subtracting the mean of $\phi$ over $\Omega$, which is an unbiased estimator of $\mu$, to obtain $\tilde{\phi}=\phi-\bar{\phi}$. We compute the conditional expectation of $\tilde{\phi}$ over $p_f(\phi)$ as
\[
\tilde{\mu}_f = E_{\phi\sim p_f}\left[ \phi - \bar{\phi}\right]  =  \mu_f - \mu = \mu_f - \left(P_f\mu_f + P_b\mu_b\right)
= \left(1-P_f\right)\mu_f - P_b\mu_b = P_b \left(\mu_f - \mu_b\right)
\]
Similarly, the conditional expectation of $\tilde{\phi}$ over $p_b(\phi)$ is 
\[
\tilde{\mu}_b= E_{\phi\sim p_b}\left[ \phi - \bar{\phi}\right]  =  \mu_b - \mu = \mu_b - \left(P_f\mu_f + P_b\mu_b\right)
= \left(1-P_b\right)\mu_b - P_f\mu_f = P_f \left(\mu_b - \mu_f\right)
\]
Now consider the neighborhood of pixels ${\cal N}(\mathbf{x})$. Without loss of generality, we will assume that the central pixel $\mathbf{x}$ is part of the foreground object. Assuming that $\phi(\mathbf{x})$ and $\phi(\mathbf{x}+\mathbf{v})$ are independently drawn from $p_f(\phi)$, we can compute the expectation of the numerator of (\ref{eqn:nsl}) as
\begin{eqnarray}
E_{\phi(\mathbf{x}),\phi(\mathbf{x}+\mathbf{v})\sim p_f}
\left[ \left\langle\tilde{\phi}(\mathbf{x}+\mathbf{v}),\tilde{\phi}(\mathbf{x})\right\rangle\right]
&=&E_{\phi(\mathbf{x}+\mathbf{v})\sim p_f}\left[\tilde{\phi}(\mathbf{x}+\mathbf{v})\right]^T E_{{\phi}(\mathbf{x})\sim p_f}\left[\tilde{\phi}(\mathbf{x})\right] \nonumber\\
&=& \tilde{\mu}_f^T  \tilde{\mu}_f = P_b^2 \parallel \mu_f - \mu_b\parallel^2.
\label{eqn:fg}
\end{eqnarray}
If $\phi(\mathbf{x}+\mathbf{v})$ belongs to the background than the same expectation becomes
\begin{eqnarray}
E_{\phi(\mathbf{x})\sim p_f,\phi(\mathbf{x}+\mathbf{v})\sim p_b}
\left[ \left\langle\tilde{\phi}(\mathbf{x}+\mathbf{v}),\tilde{\phi}(\mathbf{x})\right\rangle\right]
&=&E_{\phi(\mathbf{x}+\mathbf{v})\sim p_b}\left[\tilde{\phi}(\mathbf{x}+\mathbf{v})\right]^T E_{\phi(\mathbf{x})\sim p_f}\left[\tilde{\phi}(\mathbf{x})\right] \nonumber \\
&=& \tilde{\mu}_b^T  \tilde{\mu}_f = P_fP_b(\mu_b-\mu_f)^T(\mu_f-\mu_b) \nonumber \\ 
&=& -P_f P_b \parallel \mu_f - \mu_b\parallel^2.
\label{eqn:bg}
\end{eqnarray}
Assuming $\phi$ has sufficient discriminative capacity between the foreground and background, i.e. $\mu_f \neq \mu_b$, we conclude that the output of the NSL, regardless of whether the $\phi$ function was learned from a different domain, is consistently positive when $\mathbf{x}$ and $\mathbf{x}+\mathbf{v}$ belong to the same region and consistently negative when they do not. Furthermore, if the variance of the distributions $p_f(\phi)$ and $p_b(\phi)$ are small compared to $||\mu_f - \mu_b||^2$, the denominator in (\ref{eqn:nsl}) can be approximated with $||\tilde{\mu}_f||^2$ and $(||\tilde{\mu}_f||^2\cdot ||\tilde{\mu}_b||^2)^{1/2}$ when $\mathbf{x}+\mathbf{v}$ is drawn from the foreground and background, respectively. This approximation normalizes (\ref{eqn:fg}) and (\ref{eqn:bg}) to $+1$ and $-1$, respectively, that leads to the appearance invariance property of $\psi(\mathbf{x})$.

\begin{figure}[htb]
\begin{center}
%\vspace{-0.5cm}
\includegraphics[width=0.98\textwidth]{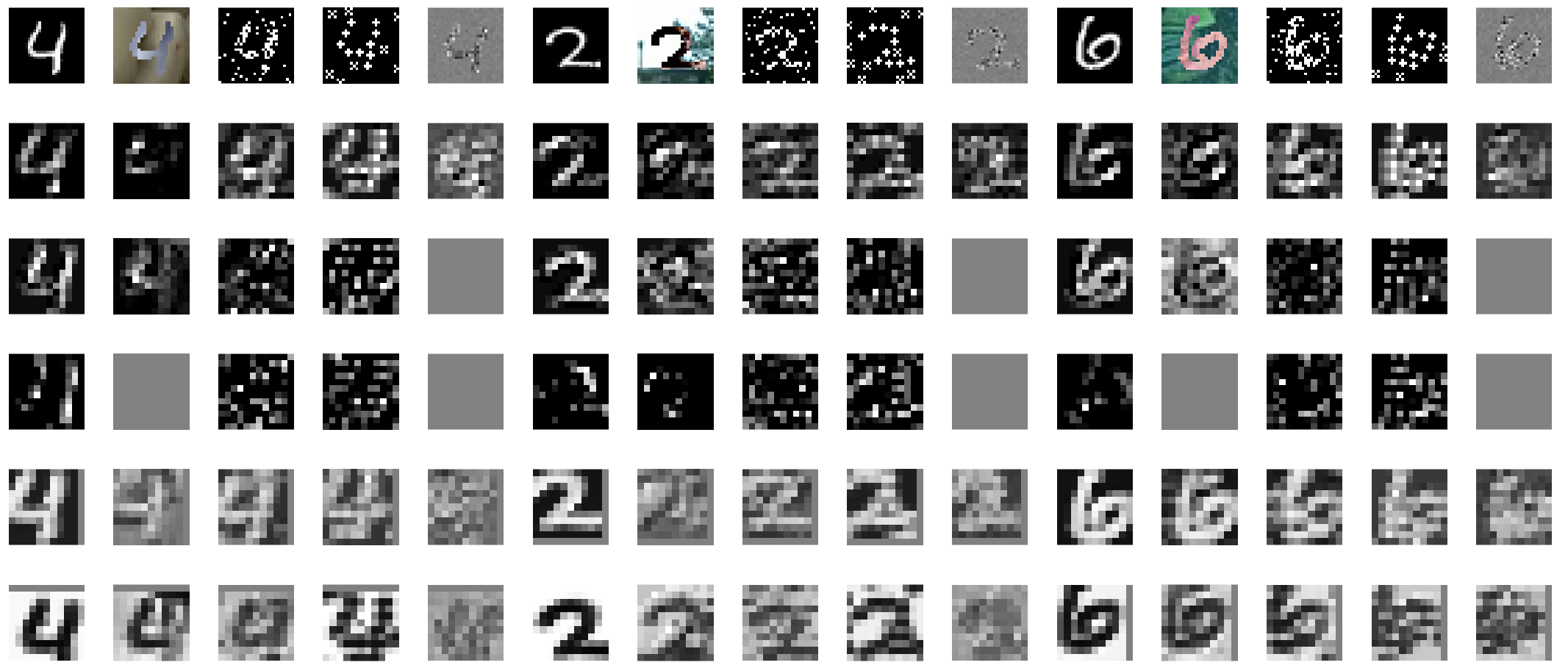}
%\vspace{-0.2cm}
\end{center}
\caption{Row 1: Images from MNIST (col. 1, 6 \& 11), MNIST-m (col. 2, 7, \& 12), and additional variants MNIST-p (col. 3, 8, \& 13), MNIST-s (col. 4, 9 \& 14), and MNIST-v (col. 5, 10 \& 15). Rows 2-4: Three feature maps from $\phi$ after first pooling layer of a network trained on MNIST images only. Rows 5-6: Neighborhood similarities computed by NSL of the same network for $\mathbf{x}$ in the foreground and in the background, respectively. Neighborhood similarities $\psi(\mathbf{x})$ exhibit increased invariance to appearance of the images in row 1 compared to the feature maps.}
\label{fig:maps}
%\vspace{-0.7cm}
\end{figure}

Figure~\ref{fig:maps} visually demonstrates the appearance invariance properties of NSL. In addition to MNIST-m~\cite{Ganin2016}  which is commonly used in domain adaptation experiments, we created three additional variants of MNIST. An MNIST image is modified by first defining a foreground by thresholding as in MNIST-m~\cite{Ganin2016}. Then, for MNIST-p, pixels in the foreground and background are set to $1$ with probabilities of $0.5$ and $0.05$, respectively. For MNIST-s, we insert a $3\times 3$ binary shape for a subset of the foreground pixels. A different shape is inserted as a distractor for a subset of the background pixels. For MNIST-v, the foreground and the background intensities are drawn from Gaussian distributions with the same mean but different standard deviations. The network with the NSL trained on MNIST (see Section~\ref{sec:digit} for details) was tested on MNIST and the MNIST-m,p,s,v variants. Figure~\ref{fig:maps} rows 2-4 demonstrate that the statistics of the feature maps in the target domains (MNIST-m,p,s,v) differ significantly from their statistics in the source domain (MNIST). On the other hand, the statistics of the similarity neighborhoods (Figure~\ref{fig:maps} rows 5-6) are consistent across the source and target domains.

\section{Experiments}

\subsection{Cross-domain Generalization in Digit Recognition}
\label{sec:digit}

{\bf Network architecture: }  We use a convolutional network with two convolutional and three fully-connected layers (Figure~\ref{fig:arch}). The first layer uses $5\times 5$ convolutions to produce $121$ feature maps followed by ReLU activation functions. This is followed by a max-pooling layer of size $2 \times 2$ and stride $2$ and a NSL with a $11\times 11$ square ${\cal N}$ producing $121$ similarity feature maps ($3\times 3$ ${\cal N}$ shown in Figure~\ref{fig:arch}). The output of NSL is passed to the second convolutional layer which produces $48$ feature maps using $5\times 5$ convolutions across $121$ channels. This is followed by ReLU activation functions, a max-pooling layer with size of $2\times2$ and stride $2$ and by two fully-connected layer with $100$ neurons and ReLU activation functions. The output layer has $10$ neurons followed by a softmax. We create a second model by removing the NSL. The number of feature maps of the first convolutional layer was chosen to match the number of maps in the output of the NSL ensuring that the input to the second convolutional layer has the same dimensionality whether NSL is included in the architecture or not. Considering that NSL is parameter free, the degrees of freedom in the model is the same with or without NSL which allows a fair comparison.  

{\bf Training/testing:}  We use Xavier initialization~\cite{Glorot2010} and use multinomial logistic loss for training. The training batch size is $64$ and the only data pre-processing is to divide each input by $256$ so that the input range is between $0$ and $1$. We train the network for $20$ epochs with an initial learning rate of $0.01$ which exponentially decays in later epochs. We train on MNIST~\cite{Lecun1998}  and test on MNIST, MNIST-m and SVHN~\cite{Netzer2011}. We also created additional variants of MNIST which we name MNIST-p, MNIST-s and MNIST-v as described in Section~\ref{sec:invariance}. The network trained on MNIST was also tested on these variants. Next we trained on SVHN and tested on SVHN and MNIST. We repeat the training of each network five times and report the mean and standard deviation of testing accuracy in Table~\ref{tab1}.  We also used domain adversarial training~\cite{Ganin2016} to adapt each network to the target domain and report another set of testing accuracies for this case.

\begin{table}[htb]
\caption{Mean $\%$ accuracy $\pm$ $\%$ standard deviation of models trained on source and tested on target (source$\rightarrow$ target). Boldface indicates state-of-the-art results (only for MNIST, MNIST-m and SVHN).}
\begin{center}
\begin{tabular}{ l | c | c | c | c | c | c | c |}
  & w/o NSL & NSL & w/o NSL+DA & NSL+DA & CKN-PM1\\
  \hline
  \scriptsize{MNIST $\rightarrow$ MNIST} & 99.22 $\pm$ 0.03 & 99.21 $\pm$ 0.05 & -- & -- & 99.43 \\
  \scriptsize{MNIST $\rightarrow$ MNIST-m}  & 57.59 $\pm$ 2.00 & {\bf 88.29} $\pm$ 0.49 & 64.25 $\pm$ 0.88 & 95.66 $\pm$ 0.32 & 46.39 \\
  \scriptsize{MNIST $\rightarrow$ SVHN}  & 32.38 $\pm$ 2.20 & {\bf 63.88} $\pm$ 1.14 & 30.22 $\pm$ 6.41  &   {\bf 64.06} $\pm$ 2.26 & 35.99 \\
 \hline
  \scriptsize{MNIST $\rightarrow$ MNIST-p} & 66.15 $\pm$ 1.28 & 81.87 $\pm$ 1.41 & 84.95 $\pm$ 0.77 & 88.16 $\pm$ 0.88 & \\
  \scriptsize{MNIST $\rightarrow$ MNIST-s} & 36.86 $\pm$ 2.19 & 55.95 $\pm$ 1.20 & 74.54 $\pm$ 1.79 & 80.32 $\pm$ 3.72& \\
  \scriptsize{MNIST $\rightarrow$ MNIST-v} & 12.03 $\pm$ 1.24 & 51.02 $\pm$ 1.39 & 11.77 $\pm$ 2.13 & 94.18 $\pm$ 0.24& \\
 \hline
 \hline
  \scriptsize{SVHN  $\rightarrow$ SVHN} & 89.10 $\pm$ 0.23 & 91.90 $\pm$ 0.20 & -- & -- & \\
  \scriptsize{SVHN  $\rightarrow$ MNIST} & 62.22 $\pm$ 2.40 & {\bf 69.77} $\pm$ 1.76 & 61.33 $\pm$ 3.37  & 74.46 $\pm$ 1.04 &  
\end{tabular}  
\end{center}
\label{tab1}
%\vspace{-0.5cm}
\end{table}

{\bf Results:} First, we observe that the inclusion of the NSL has no detrimental effect on the accuracy of the network for MNIST $\rightarrow$ MNIST, and it improves the accuracy by a modest amount ($2.8\%$) for SVHN  $\rightarrow$ SVHN. When the source and target domains differ, NSL provides a large increase to accuracy in all cases. All of the results with NSL are state-of-the-art results surpassing source only experiments in the domain adaptation literature which is the proper comparison. Furthermore, the state-of-the-art domain adaptation result for MNIST$\rightarrow$SVHN is 52.8\%~\cite{Saito2017} to the best of our knowledge. The network with NSL without using any domain adaptation achieves $63.88\%$. For MNIST-v, the network without NSL has accuracy around chance level ($10\%$) whereas the network with NSL is $51\%$ accurate. We note that using domain adversarial training with NSL further increases accuracy in all cases. For MNIST-v, domain adversarial training without NSL provides no gains above chance level whereas with NSL the accuracy is $94.18\%$.
The accuracy when using domain adversarial training and NSL together for SVHN$\rightarrow$MNIST is $74.46\%$. Previous domain adversarial results for this task is $73.85\%$~\cite{Ganin2016} while the state-of-the-art result for this task is $86.2\%$ which uses tri-training~\cite{Saito2017}. This suggests that using tri-training with NSL could lead to state-of-the-art results for the SVHN$\rightarrow$MNIST task in the future. Finally, we also experimented with CKN~\cite{Mairal2014}. We tried the patch and gradient maps for CKN and found CKN-PM1~\cite{Mairal2014} to give the best results in the target domain which are reported in Table~\ref{tab1}. While good results were obtained for MNIST $\rightarrow$ MNIST, the trained CKN was not found to generalize well to MNIST-m and SVHN. 

\subsection{Object Recognition}

While appearance should have no role in digit classification, this is not necessarily the case for general object recognition. We use a simple convolutional network to study the impact of NSL on object recognition with the CIFAR-10 dataset~\cite{krizhevsky2009learning}. 

{\bf Network architecture: } The base network consists of 3 convolutional layers, each with $5\times5$ kernels (32, 32 and 64 maps) followed by ReLU activations and $3\times 3$ pooling with stride 2. This is followed by a fully connected layer with 64 neurons and an output layer with 10 neurons (softmax).  This network is trained on the CIFAR-10 dataset using multinomial logistic loss. We also modify the base network to insert a NSL with a $15\times 15$ ${\cal N}$ between the first pooling and second convolutional layers. 

{\bf Training/testing:}   We perform preprocessing by subtracting the mean image of the training set from each image. We do not perform any type of data augmentation. Weights are initialized randomly from a Gaussian distribution and trained for 20 epochs. The learning rate is $0.001$ for the first $16$ epochs and $0.0001$ for the last $4$ epochs. We trained both networks five times. 
   
 {\bf Results:} The mean $\pm$ standard deviation testing accuracies for the base network and base+NSL are $76.60 \pm 0.42$ and $66.36\pm 0.70$, respectively. We attribute this decrease in accuracy to appearance information that is relevant to the CIFAR classification task that is discarded by NSL. 
  
{\bf Modified network architecture: }  To remedy the decrease in accuracy, we create a new network by inserting a $25$ channel network in network layer (NiN) between the first pooling and second convolutional layers, in parallel to the NSL. The outputs of NSL and NiN are concatenated before being passed to the second convolutional layer. The testing accuracy of this network is $79.09\pm 0.22$ which surpasses the base network accuracy. If the NiN is included but NSL is excluded the accuracy is $77.07 \pm 0.25$ demonstrating that both NSL and NiN contribute to improved accuracy. 

The final question that remains is whether the addition of a parallel NiN stream to NSL impacts generalization accuracy when the training and testing domains differ. We use the same network described in Section~\ref{sec:digit} for digit recognition and add a NiN with 25 feature maps after the first pooling layer. Then, we concatenate the output of NiN with the output of NSL and feed it to the second convolutional layer. The learning rate and the number of epochs are the same as the experiments of Section~\ref{sec:digit}. The accuracy of this network for MNIST $\rightarrow$ MNIST-m is $87.70\pm 0.20$ demonstrating that the parallel NiN stream does not adversely affect generalization when the training and testing domains differ. One potential explanation is that the parallel NiN and NSL streams decouple appearance and shape, and the following layers learn to use the relevant information. 

\subsection{Cross-domain Vasculature Segmentation}

\begin{figure}[htb]
\begin{center}
\begin{tabular}{cccc}
{\includegraphics[width = 0.21\textwidth]{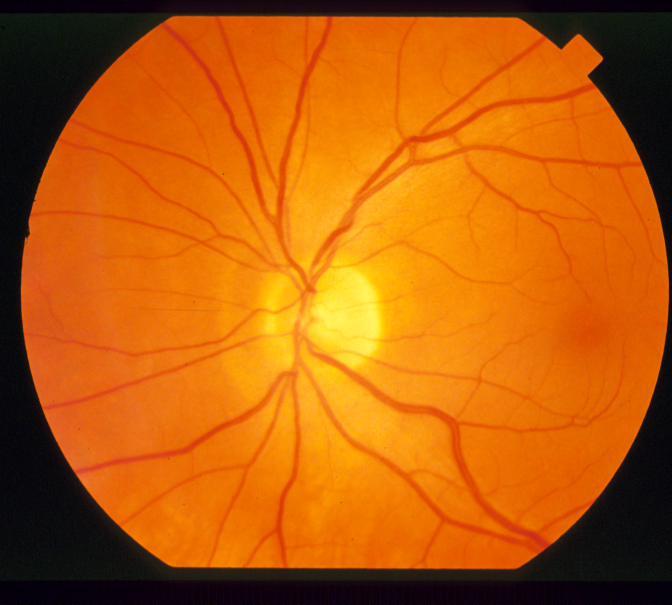}} &
{\includegraphics[width = 0.21\textwidth]{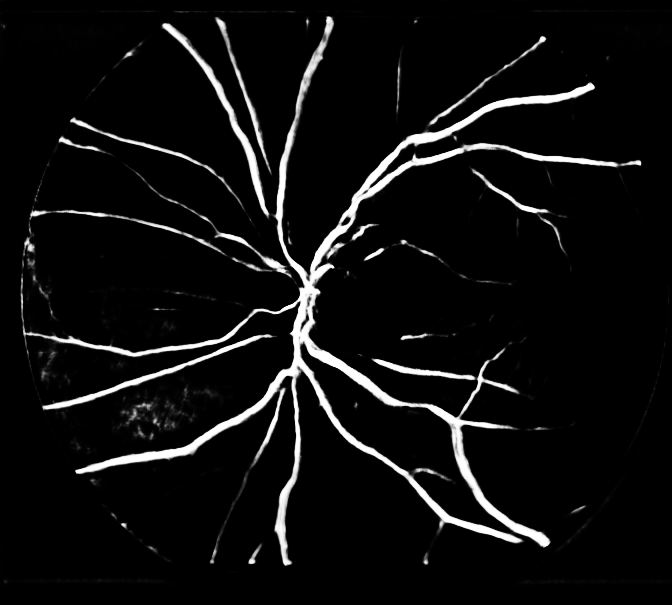}} &
{\includegraphics[width = 0.21\textwidth]{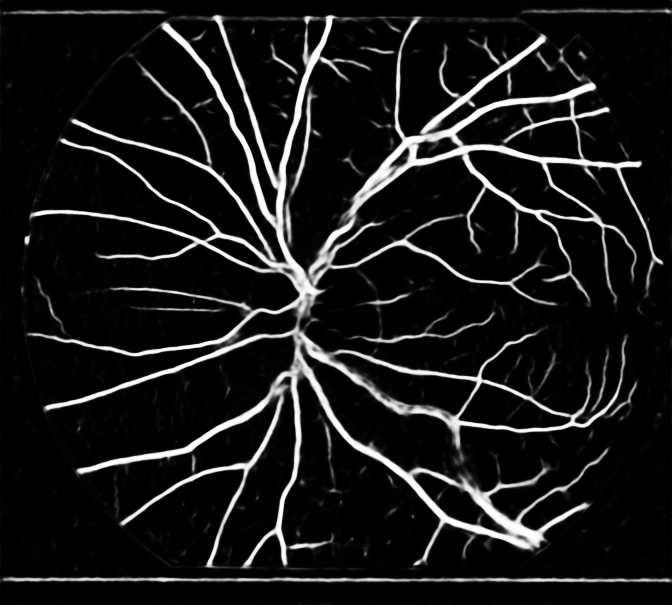}} &
{\includegraphics[width = 0.21\textwidth]{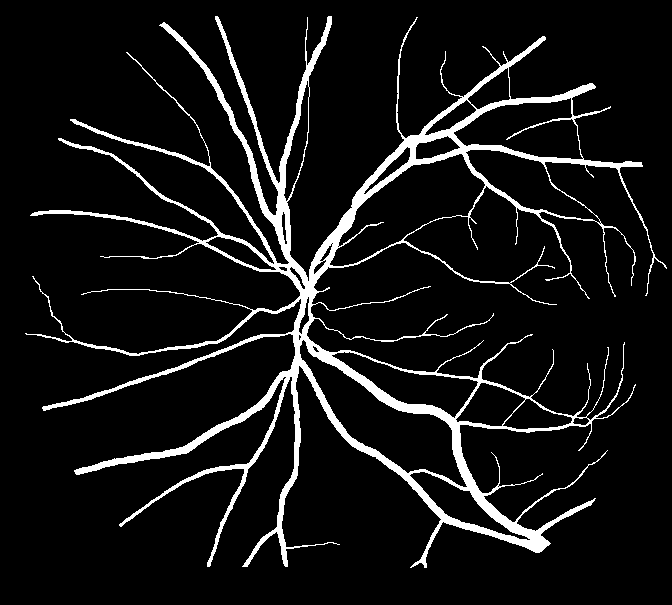}}\\
{\includegraphics[width = 0.21\textwidth]{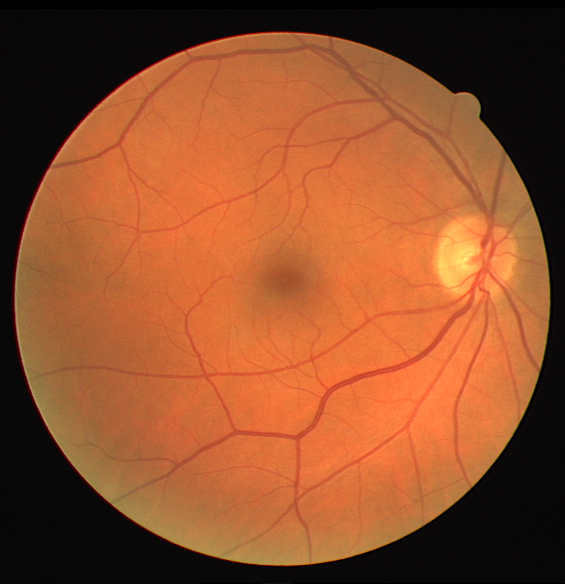}} &
{\includegraphics[width = 0.21\textwidth]{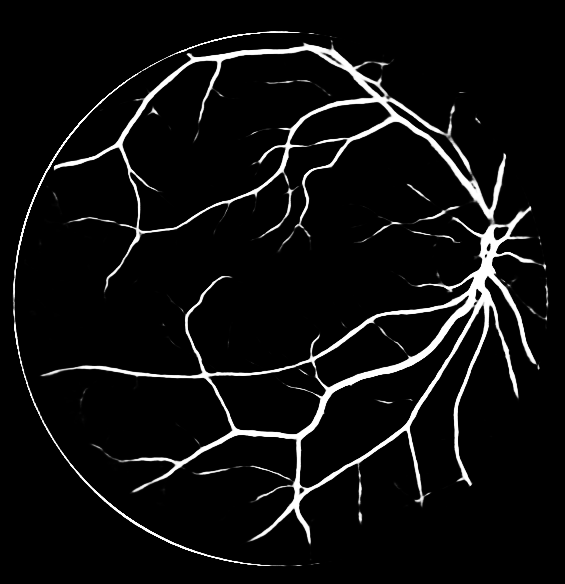}} &
{\includegraphics[width = 0.21\textwidth]{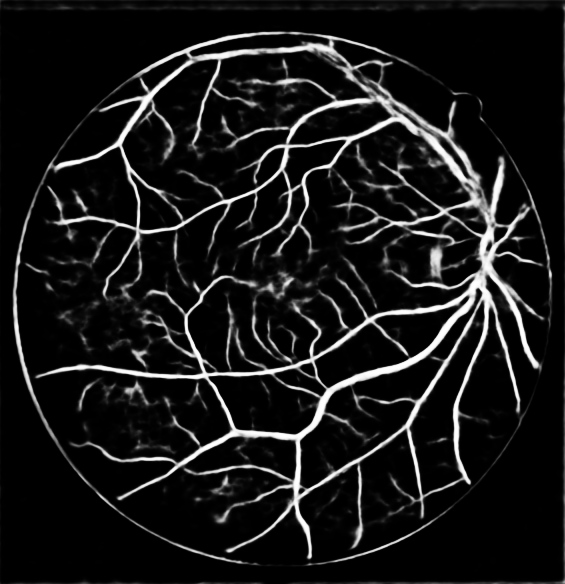}} &
{\includegraphics[width = 0.21\textwidth]{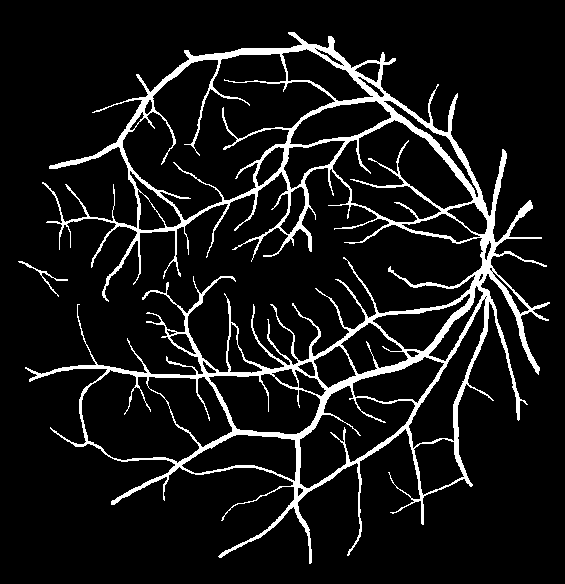}}\\
\end{tabular}
\end{center}
\caption{Top row: DRIVE $\rightarrow$ STARE; bottom row: STARE $\rightarrow$ DRIVE. Left to right: Fundus image from test set, without NSL, with $11\times 11$ NSL, expert ground truth.\label{fig:eye}}
\end{figure}

\begin{table}
%\vspace{-0.7cm}
  \caption{Test set F-score of models trained on source and tested on target (source$\rightarrow$ target).}
  \begin{center}
  \begin{tabular}{ c | c | c | c }
    & w/o NSL  & 5$\times$5 ${\cal N}$  &  11$\times$11 ${\cal N}$ \\
    \hline
       DRIVE $\rightarrow$ DRIVE    &   80.34   &  80.41   &   78.83 \\
          STARE $\rightarrow$ DRIVE  &   62.22 & 69.93 &  69.92 \\         
   \hline
      STARE $\rightarrow$ STARE    &   73.13   &  78.08  &   78.78 \\     
         DRIVE $\rightarrow$ STARE    &   65.20   &  70.49   &   74.00 \\
  \end{tabular}  
  \end{center}
  \label{tab2}
\end{table}
We evaluated the proposed NSL for vasculature segmentation in eye fundus images using the U-net architecture~\cite{Ronneberger2015}, a fully convolutional network for biomedical image segmentation. For the baseline, we use the exact same network as introduced in the original paper without any modification to label pixels as vasculature or background. We also modify this network by adding a NSL after the first convolution layer. We use the DRIVE~\cite{Staal2004} and STARE~\cite{Hoover2000} datasets which contain 40 and 20 images, respectively, with manually annotated vasculature. We divide the datasets into $50\%$ training and $50\%$ testing for our experiments. We measure the accuracy of vasculature segmentation using the F-score which is defined as $2\times precision\times recall / (precision + recall)$. Table~\ref{tab2} shows testing accuracies for different combinations of source and target datasets as well as different sizes of ${\cal N}$ for NSL.  First, we observe that the inclusion of NSL increases the accuracy of segmentation for STARE $\rightarrow$ STARE while moderately decreasing it for DRIVE $\rightarrow$ DRIVE. More importantly, the inclusion of NSL significantly increases the accuracy of the segmentation when the target and source domain differ. Figure~\ref{fig:eye} illustrates that the recall of vasculature pixels is significantly improved in cross domain experiments with the inclusion of the NSL. 
We emphasize that no domain adaptation is used for these experiments. A larger neighborhood size gives more accurate results for DRIVE $\rightarrow$ STARE, whereas the results for STARE $\rightarrow$ DRIVE do not significantly differ between $5\times 5$ and $11\times 11$ ${\cal N}$. Remarkably, the results for DRIVE $\rightarrow$ STARE with a $11\times 11$ ${\cal N}$ surpass the results for STARE $\rightarrow$ STARE without NSL. To demonstrate that the same improvements in accuracy can not be obtained by histogram matching between domains, we preprocessed both source and target domain images by converting to hue, saturation and intensity space, performing histogram equalization on the intensity values, and converting back to RGB. The network without NSL was found to have F-scores of $79.10\%$ and $59.91\%$ for DRIVE $\rightarrow$ DRIVE  and DRIVE $\rightarrow$ STARE, respectively. Finally, we note that the state-of-the-art results for vasculature segmentation are obtained with a convolutional network that includes specialized side layers~\cite{Maninis16}. Our NSL layer can be incorporated into such specialized networks to improve their performance and cross-domain applicability further. 
 
\subsection{Cell Detection}
\begin{figure}[htb]
\begin{center}
\begin{tabular}{cccc}
\includegraphics[width=0.21\textwidth]{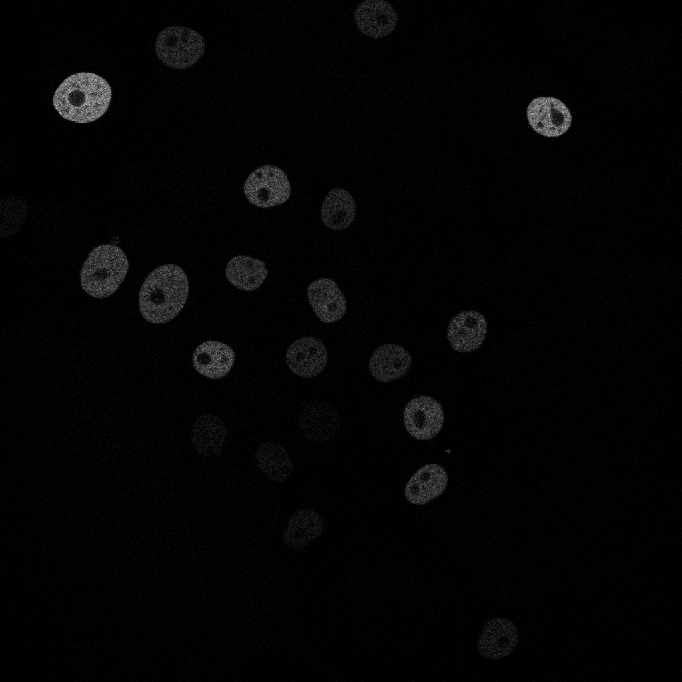}&
\includegraphics[width=0.21\textwidth]{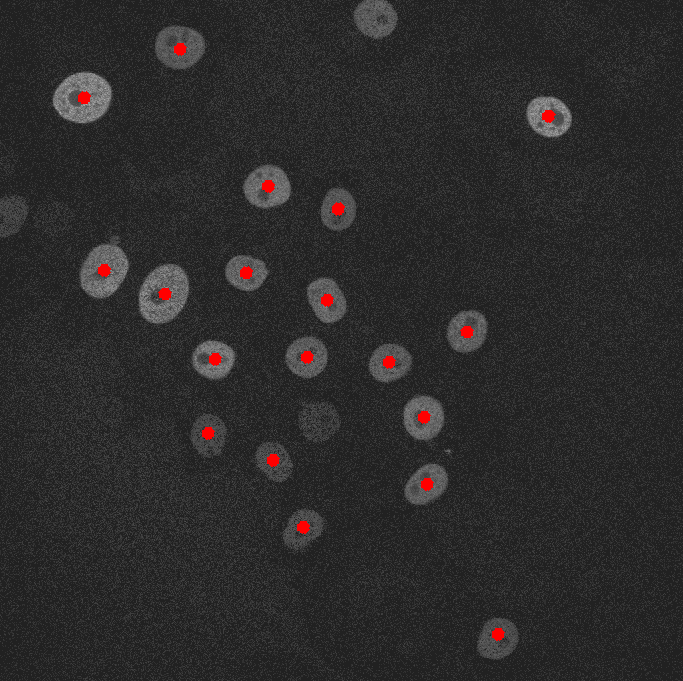}&
\includegraphics[width=0.21\textwidth]{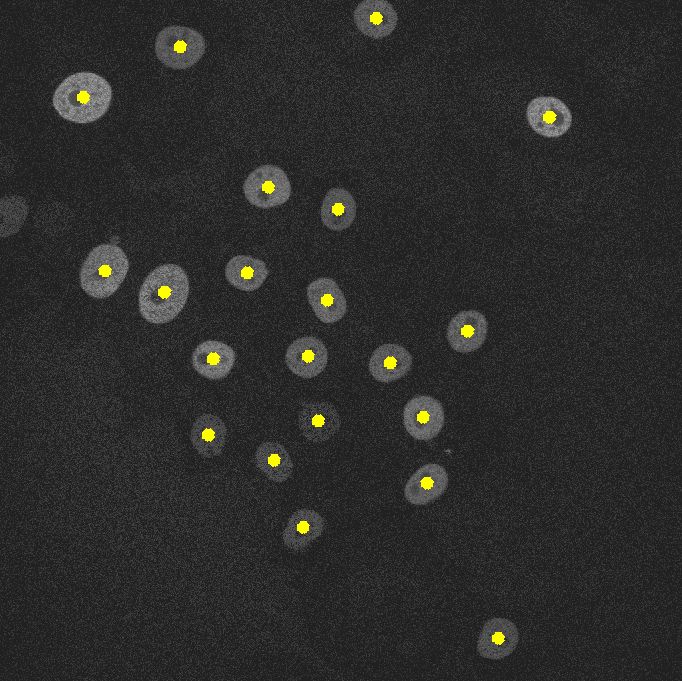}&
\includegraphics[width=0.21\textwidth]{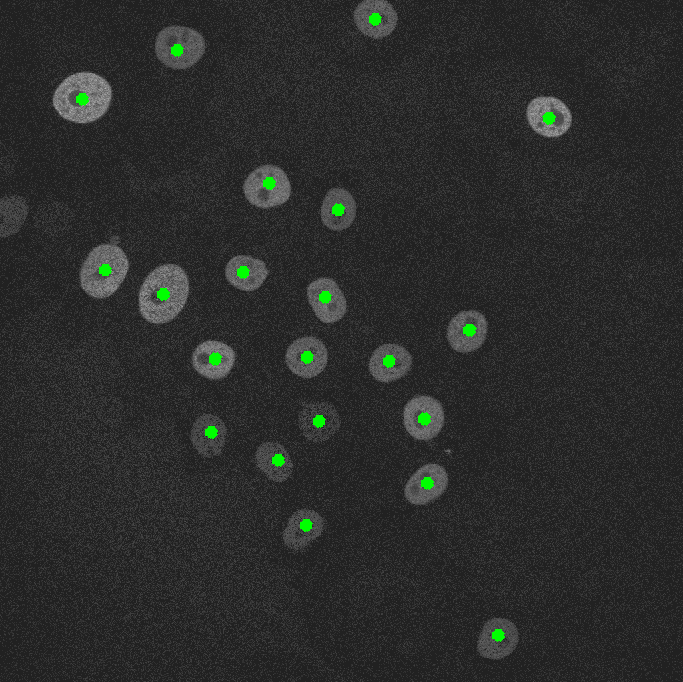}\\
\includegraphics[width=0.21\textwidth]{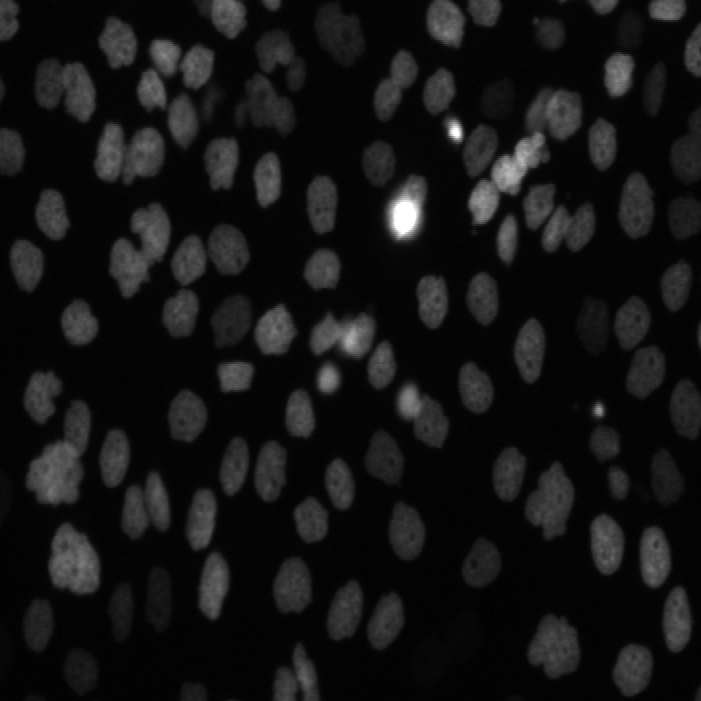}&
\includegraphics[width=0.21\textwidth]{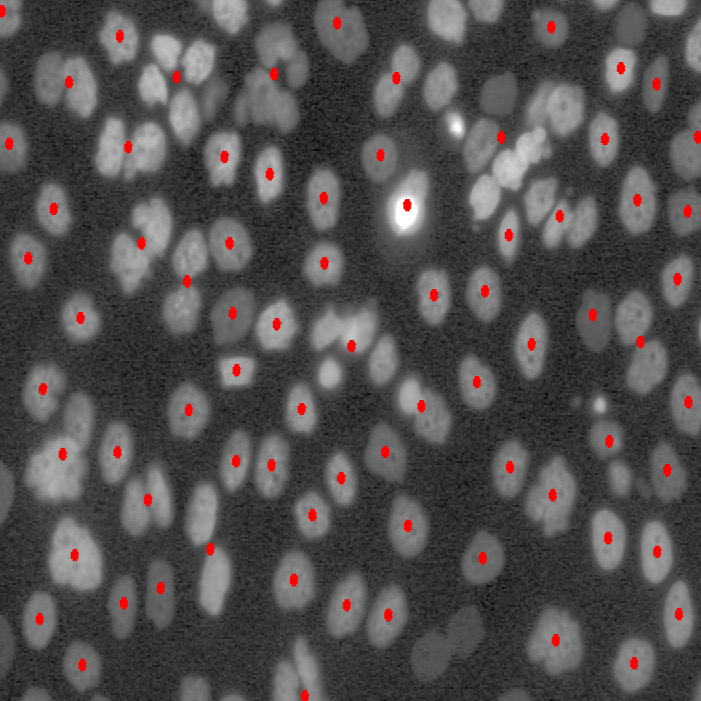}&
\includegraphics[width=0.21\textwidth]{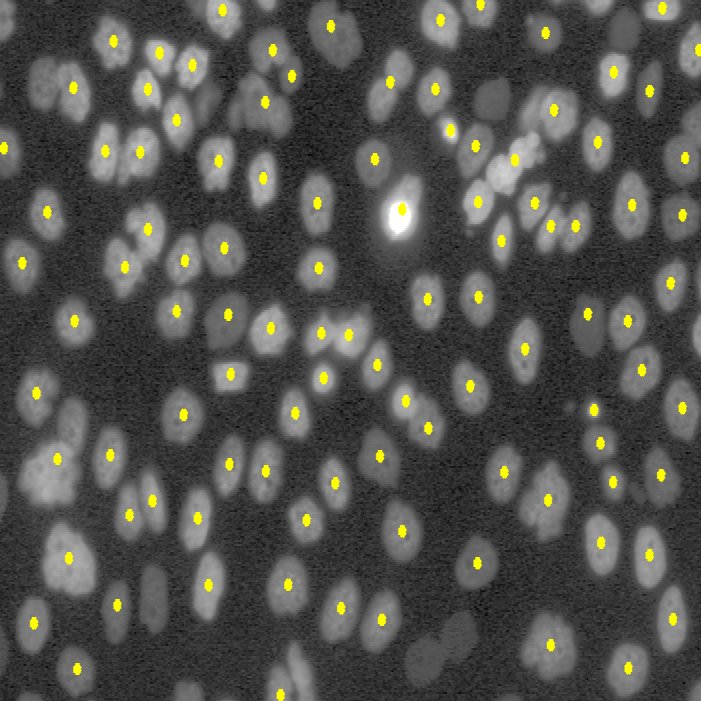}&
\includegraphics[width=0.21\textwidth]{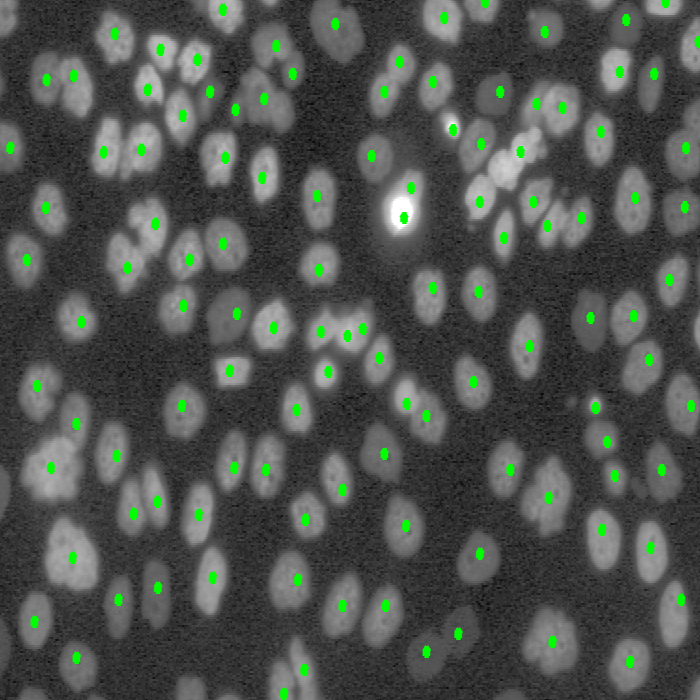}
\end{tabular}
\end{center}
\caption{Top: Fluo-N2DH-GOWT1-1; Bottom: Fluo-N2DL-HeLa-2~\cite{Solorzano}. Left to right: Test image, predicted centroids without NSL, predicted centroids with NSL and ground truth centroids overlaid on contrast enhanced images (for display only). }
\label{fig:cells}
%\vspace{-0.5cm}
\end{figure}

We used time sequences from two datasets from the cell tracking challenge~\cite{Solorzano}. Both datasets have two sequences each, one for training and one for testing. The ground truth for these datasets consists of dot annotations at approximately the centers of cells. Our goal is to detect cells, i.e. cell centers; therefore, we place small Gaussian masks centered at the dot annotations to create the target output images for training. We train the U-net architecture~\cite{Ronneberger2015} and a modified U-net where we add a NSL with $9\times 9$ ${\cal N}$ after the first convolution layer. Visual results for cell detection on both the datasets are seen in Figure \ref{fig:cells}.  With the addition of NSL, we are able to identify cells which have low contrast in the Fluo-N2DH-GOWT1 dataset. In the case of the Fluo-N2DL-HeLa dataset, we observe that some of the cells that have been detected as clumps are individually identified with the NSL. For quantitative evaluation, we use the Hungarian algorithm~\cite{Kuhn1955} to match ground truth dot annotations with local maxima of the network output and compute precision, recall, and F-score. We have also compared our detection results with state-of-the-art results from~\cite{Turetken2017} and evaluated the U-net without NSL using locally contrast enhanced images. The quantitative results from our evaluations are reported in Table \ref{tab:results}. We note that the improvement with NSL significantly exceeds the improvement obtained with contrast enhancement of the input images and provides state-of-the-art accuracy in terms of F-score.  
\begin{table}[htb]
\caption{Cell detection testing accuracy.}
\label{tab:results}
\begin{center}
\begin{tabular}{c || c | c | c || c | c | c|| }
Data set & \multicolumn{3}{c||}{Fluo-N2DH-GOWT1-1} & \multicolumn{3}{c||}{Fluo-N2DL-HeLa-2 }\\
Method & Precision & Recall & F-score & Precision & Recall & F-score \\ 
\hline
ECLIP \cite{Turetken2017} & {\bf 100.0} & 31.0 & 47.33 & 78.0 & {\bf 86.0} & 81.80\\ 
 U-net  &  94.91 & 82.71 & 88.39 & 85.74 & 69.94 & 77.03 \\ 
U-net + contrast enhanced & 94.66 & 86.16 & 90.21 & 85.86 & 71.97 & 78.30\\ 
U-net + $9\times 9$ NSL & 96.95 & {\bf 94.59} & {\bf 95.75} & {\bf 89.60} & 79.49 & {\bf 84.24}\\ 
\end{tabular}
\end{center}
\end{table}

%\vspace{-0.5cm}

\section{Conclusion}
We proposed a NSL that transforms its input feature map using the feature vectors at each pixel as a frame of reference for its surrounding neighborhood. This spatial adaptation was shown to induce appearance invariance in a network when used in conjunction with convolutional layers. NSL is a parameter free layer that can be included after any image feature map layer in a network. NSL is differentiable; therefore, networks including a NSL can be trained in an end-to-end manner. We analyzed invariance properties of NSL and demonstrated in the context of digit recognition and biomedical image segmentation that the inclusion of a NSL allows networks to better generalize to data that are not well represented by the training data without requiring domain adaptation. We showed that the inclusion of NSL also provides a further increase to accuracy for domain adaptation. Future directions for research include NSL used after deeper layers of a network and the use of a NiN and NSL to decouple appearance and geometry for general object recognition. Specific domain adaptation techniques for networks with NSL/NiN combination are also of interest. Finally, we plan to investigate further applications of NSL in biomedical image analysis applications where labeled training data might not be readily available.

\newpage
\bibliographystyle{ieeetr}
\bibliography{nips2017}

\end{document}